\documentclass[10pt,twocolumn,letterpaper]{article}

\usepackage[table]{xcolor}
\usepackage{iccv} 
\usepackage{times}
\usepackage{epsfig}
\usepackage{graphicx}
\usepackage{amsmath}
\usepackage{amssymb}
\usepackage{cite}
\usepackage{multirow}
\usepackage{rotating}
\usepackage{pifont}
\usepackage{array,arydshln}
\usepackage{commath}
\usepackage{verbatim}

\iftrue
\newcommand{\davide}[1]{\textcolor{blue}{\bf [DAVIDE: #1]}}
\newcommand{\rahul}[1]{\textcolor{green}{\bf [RAHUL: #1]}}
\newcommand{\bing}[1]{\textcolor{orange}{\bf [BING: #1]}}
\newcommand{\joe}[1]{\textcolor{magenta}{\bf [JOE: #1]}}
\newcommand{\todo}[1]{\textcolor{red}{\bf [TODO: #1]}}

\else
\newcommand{\davide}[1]{}
\newcommand{\rahul}[1]{}
\newcommand{\bing}[1]{}
\newcommand{\joe}[1]{}
\newcommand{\todo}[1]{}
\fi

\newcommand{\cmark}{\ding{51}}

\usepackage[pagebackref=true,breaklinks=true,letterpaper=true,colorlinks,bookmarks=false]{hyperref}

\DeclareMathOperator*{\mm}{\mathbf{M}}

\iccvfinalcopy 
\def\iccvPaperID{1346} 

\ificcvfinal\fi
\begin{document}

\title{Multi-Scale Attention Network for Crowd Counting}

\author{Rahul Rama Varior, Bing Shuai, Joseph Tighe, Davide Modolo\\
Amazon AI\\
{\tt\small rahulrv,bshuai,tighej,dmodolo@amazon.com}
\\
}

\maketitle

\vspace{-6mm}
\begin{abstract}
\vspace{-3mm}
In crowd counting datasets, people appear at  different scales, depending on their distance from the camera. To address this issue, we propose a novel multi-branch scale-aware attention network that exploits the hierarchical structure of convolutional neural networks and generates, in a single forward pass, multi-scale density predictions from different layers of the architecture.  To aggregate these maps into our final prediction, we present a new soft attention mechanism that learns a set of gating masks.  Furthermore, we introduce a scale-aware loss function to regularize the training of different branches and guide them to specialize on a particular scale. As this new training requires annotations for the size of each head, we also propose a simple, yet effective technique to estimate them automatically. Finally, we present an ablation study on each of these components and compare our approach against the literature on 4 crowd counting datasets: UCF-QNRF, ShanghaiTech A \& B and UCF\_CC\_50. Our approach achieves state-of-the-art on all them with a remarkable improvement on UCF-QNRF (+$25$\% reduction in error).

\end{abstract}

\vspace{-5mm}
\section{Introduction}
\vspace{-1mm}
Crowd counting is the task of predicting the number of people present in an image and in recent years, it has attracted growing interest in the academic research community. 
The computer vision community has tackled this task in a variety of ways: early works either counted based on the outputs of a body or head detector~\cite{wu05iccv,wang11cvpr,rodriguez11iccv} or learned a mapping from the global or local features of an image to the predicted count~\cite{chan08cvpr,chan09cvpr,ryan09dicta}. More recently, thanks to the ability of convolutional neural networks to learn local patterns, works have started to learn density maps that not only predict the count, but also the spatial extent of the crowd~\cite{zhang16cvpr, boominathan16acm, onoro16eccv, sam17cvpr, sindagi17iccv, zhang18wacv, sam18cvpr, li18cvpr, liu18cvpr, shen2018cvpr, idrees18eccv, cao18eccv}.

Despite this progress, crowd counting remains a challenging task due to background clutter, heavy occlusions and scale variations. Of these, scale is the issue that has received the largest amount of attention in recent literature~\cite{zhang16cvpr, sam17cvpr, sam18cvpr,sindagi17iccv,onoro16eccv,boominathan16acm,zhang18wacv, cao18eccv,li18cvpr}.

 \begin{figure}
  \begin{center}
    \includegraphics[width=0.45\textwidth]{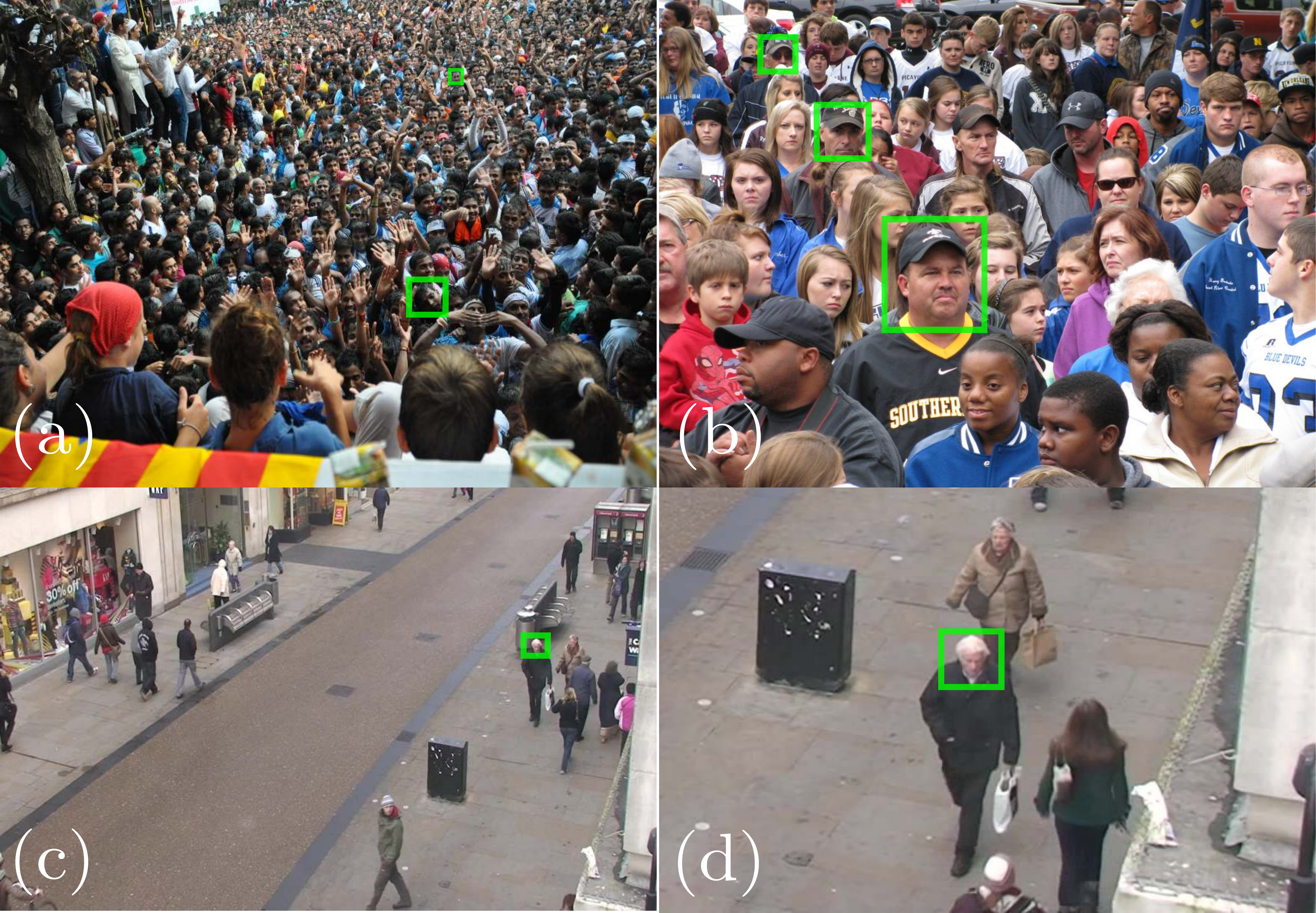}
\end{center}
\vspace{-3mm}
\caption{\small \it 
One of the most impactful issue in crowd counting is scale variation. For example, two similar people appear visually different when they are not at the same distance from the camera (a-b); and the same person can appear small in an image, but much larger in another (c-d). We tackle the former case with a novel scale-aware convolutional neural network, while the latter with a simple, yet effective image size regularization approach.\vspace{-3mm}}
\label{fig:scale_intro}
\end{figure}

In this paper, we tackle the notion of scale that deals with visual changes in people's appearance with respect to their distance from the camera. As pictured in fig.~\ref{fig:scale_intro}{\color{red}a-b}, two similar individuals can appear very different depending on their relative location in the scene.  
To solve this issue, we propose a novel scale-aware deep convolutional neural network. 
%
The hierarchical structure of convolutional neural networks progressively expands the receptive field of the network feature maps, implicitly capturing information at different scales.
Inspired by the skip branches in FCN~\cite{long15cvpr} and SSD~\cite{liu16eccv}, we propose to directly generate multiple density maps from these intermediate feature maps.
%
As the feature map generated by the last convolutional layer has the largest receptive field, it carries high-level semantic information that can be used to localize {\it large} heads; on the other hand, feature maps generated by the intermediate layers are more accurate and robust in counting extremely {\it small} heads (i.e., the crowds), and they contain important details about the spatial layout of the people and low-level texture patterns.

In order to aggregate the density maps generated from different layers of our network, we propose a novel {\it soft attention mechanism} that learns a set of gating masks, one for each map. Our masks learn to attend to large heads from the density map predicted by the last convolutional layer and smaller ones from earlier layers.
While this can be trained by only providing supervision to the final density estimate, we found that performance improves by supervising the intermediate density estimates as well. We propose a new {\it scale-aware loss function} to further regularize our multi-scale estimates and guide them to specialize on a particular head size. Furthermore, as head size information is not available in any crowd counting dataset, we also propose a novel approach to automatically estimate them. Our approach combines the geometry-adaptive technique of ~\cite{zhang16cvpr} with a new bounding-box-adaptive technique.

In our experiments we show that our approach achieves state-of-the-art results on four major crowd counting datasets: UCF-QNRF~\cite{idrees18eccv}, ShanghaiTech A \& B~\cite{zhang16cvpr} and UCF\_CC\_50~\cite{idrees13cvpr}, with a substantial improvement on UCF-QNRF (over $25$\% reduction in error). Moreover, in our ablation study we analyze the density maps generated by different layers of our network and show that each specializes on the different scale variations.

To summarize, we make the following contributions:
\begin{enumerate}
\item a new network architecture that generates multi-scale density maps from its intermediate layers (sec.~\ref{sec:method});
\item a new scale-aware attention mechanism to aggregate these maps into our final prediction (sec.~\ref{sec:sel});
\item a new scale-aware loss function to further help regularize these maps during training (sec.~\ref{sec:scaleawareloss});
\item a simple, yet effective technique to estimate the size of each head in an image, in a completely automatic way (sec.~\ref{sec:scale_data_collection}).
\end{enumerate}

\section{Related work} \label{sec:rl}
\vspace{-1mm}
\paragraph{Multi-scale models for crowd counting.} 

People appear at different sizes in crowd counting images due to large perspective changes in the scenes as well as varying image resolution.  In order to address this issue, many recent works on crowd counting have focused on learning multi-scale models.

Most previous works use a multi-column architecture~\cite{zhang16cvpr, sam17cvpr, sam18cvpr,sindagi17iccv,onoro16eccv,boominathan16acm,kang18bmvc}.
Zhang et. al. trained a custom network with three CNN columns, each with a different receptive field to capture a specific range of head sizes (MCNN~\cite{zhang16cvpr}). 
Running three CNN columns was however slow and Sam et. al. proposed to predict which column to run for each input image patch (Switch-CNN~\cite{sam17cvpr}). 
Later, Sam et. al. further extended their previous work by training a mixture of experts (each one equivalent to a column) in an incrementally growing fashion (IG-CNN~\cite{sam18cvpr}). Furthermore, Sindagi et. al. proposed a new architecture where MCNN is enriched with two additional columns capturing global and local context (CP-CNN~\cite{sindagi17iccv}).
Instead of each column designed with different receptive fields, Boominathan et al. proposed using columns of different depths, where the deep CNN captured large crowds and the shallow CNN smaller ones (CrowdNet~\cite{boominathan16acm}). Finally, Onoro-Rubio et. al. (Hydra CNN~\cite{onoro16eccv}) and Kang et al. (AFS-FCN~\cite{kang18bmvc}) represented columns as pyramid levels over image patches at multiple scales (former) or over the full image fed to the same network multiple times at different resolutions (latter).
%
While all these multi-column architectures have shown promising results, they present several disadvantages: they have a large amount of model parameters, which often results in difficulties during training, and they are slow at inference, as multiple CNNs need to be run.

To overcome these limitations, recent works have focused on multi-scale, single column architectures~\cite{zhang18wacv, cao18eccv,li18cvpr}.
Zhang et. al. proposed an architecture that combines two feature maps of two layers through a skip connection (saCNN~\cite{zhang18wacv}). 
Cao et. al. proposed an encoder-decoder network, where the encoder learns scale diversity in its features by using an aggregation module that combines filters of different sizes (SANet~\cite{cao18eccv}).
Finally, Li et. al. replaced some pooling layers in the CNN with dilated convolutional filters at different rates, which enlarge the receptive field of feature maps without losing spatial resolution (CSRNet~\cite{li18cvpr}).  

In this paper, we present a single column network architecture that mimics multi-columns by predicting multi-scale density maps from different layers of the network. Our architecture takes advantage of the multi-column approaches ability to predict multi-scale density maps, yet it is much faster to compute and requires far fewer parameters. Moreover, differently from the previous multi-column approaches, our architecture aggregates its predictions using a novel attention-based mechanism that selects each column based on the size of each head in an image.

\begin{figure*}
\begin{center}
    \includegraphics[width=0.9\textwidth]{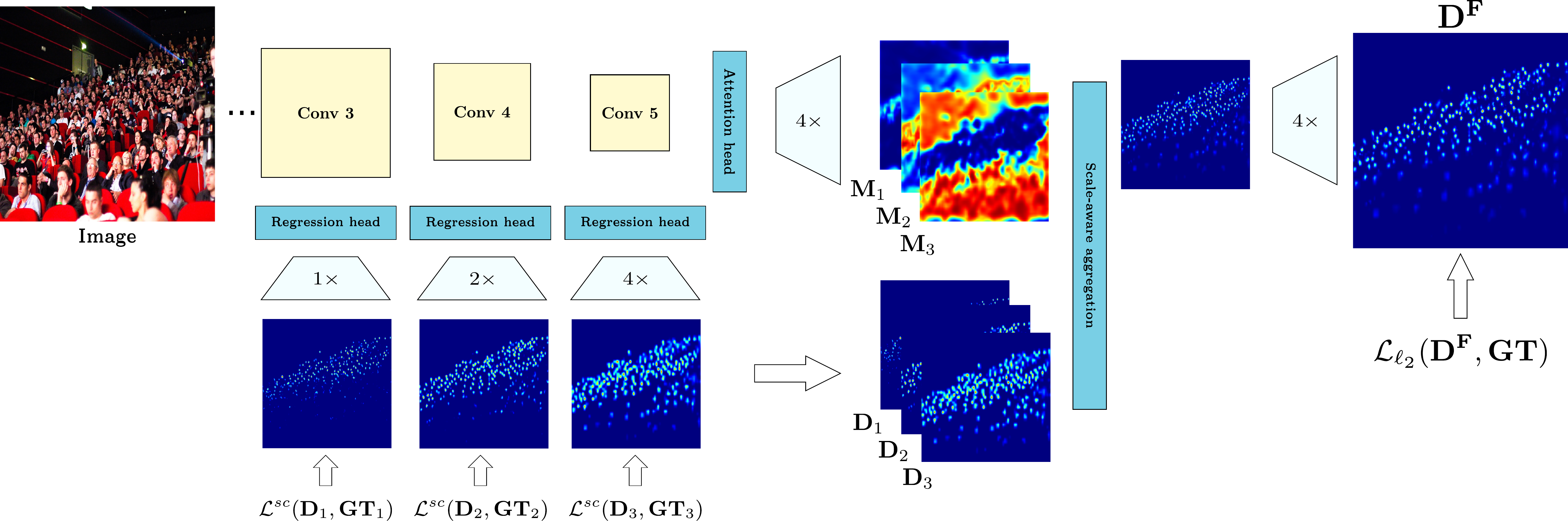}
\end{center}
\vspace{-3mm}
\caption{\small \it Our multi-branch architecture. Intermediate feature maps are used to generate density map predictions through different branches. The scale-aware attention masks $\mm_i$ are used to aggregate these maps $\mathbf{D}_i$ and generate our final prediction $\mathbf{D}^F$. Finally,  a new scale-aware loss $\mathcal{L}^{sc}$ is used to regularize the training of our branches and further help them learn different scale variations.}
\label{fig:network_architecture}
\end{figure*}

\vspace{0.5mm}
\noindent {\bf Attention-based mechanism.} Attention models have been widely used for many computer vision tasks like image classification~\cite{xiao15cvpr,cao15iccv}, object detection~\cite{ba15iclr,yoo15cvpr}, semantic segmentation~\cite{chen16cvpr,li18bmvc}, saliency detection~\cite{liu2018cvpr} and, very recently, crowd counting~\cite{kang18bmvc}.
These models work by learning an intermediate attention map that is used to select the most relevant piece of information for visual analysis. The most similar works to ours are the ones of Chen et al.~\cite{chen16cvpr} and Kang et al.~\cite{kang18bmvc}. Both approaches extract multi-scale features from several resized input images and use an attention mechanism to weight the importance of each pixel of each feature map. One clear drawback of these approaches is that their inference is slow, as each test image needs to be re-sized and fed into the CNN model multiple times. Instead, our approach is much faster: it requires a single input image and a single pass through the model, as our multi-scale features are generated by pooling information from different layers of the same network instead of multiple passes through the same network.

\section{Our approach} \label{sec:method}

In sec. \ref{sec:baseline} we present our baseline network for estimating the density map and training loss. In sec.~\ref{sec:sel} we describe how we extend this baseline with (i) our novel multi-branch density prediction architecture and (ii) our attention mechanism for selecting between these branches. In sec~\ref{sec:scaleawareloss}, we then describe our novel scale-aware loss function, which guides each density prediction branch to specialize on a particular head size. This loss requires a head size estimate during training. As head size information is not available in any public dataset for crowd counting, in sec.~\ref{sec:scale_data_collection}, we present a novel approach to automatically estimate head size.

\subsection{Baseline network for crowd counting} \label{sec:baseline}

Like other density-based approaches~\cite{zhang16cvpr, boominathan16acm, onoro16eccv, sam17cvpr, sindagi17iccv, zhang18wacv, sam18cvpr, li18cvpr, liu18cvpr, shen2018cvpr, idrees18eccv, cao18eccv} for crowd counting, given an image, we feed it to a fully convolutional network and estimate a density map (fig.~\ref{fig:network_architecture}, $\mathbf{D^F}$). Then, we sum all the values in this map to obtain the final count.

Our baseline network consists of three components: a backbone network, a regression head and an upsampling layer. 
The image is fed into the backbone, which progressively down samples the spatial resolution to produce a feature map with a large receptive field but at $1/16^{th}$ of the image resolution. These features are fed into the regression head
to produce a density map. Then bi-linear upsampling is used to bring the density estimates back to the original image resolution.

During training, we use a pixel-wise Euclidean loss on the density map output:
\begin{equation}
\mathcal{L}_{\ell_2} = \frac{1}{WH} \sum_{i}^{W} \sum_{j}^{H} \norm{\mathbf{D^F}^{(i,j)} - \mathbf{GT}^{(i,j)}}_{2}^{2}
\label{eq:l2loss}
\end{equation}
where $\mathbf{D^F}^{(i,j)}$ is the estimated density map at pixel location $(i,j)$ and $\mathbf{GT}^{(i,j)}$ is its corresponding ground truth value.
We follow the method of MCNN~\cite{zhang16cvpr} to generate the ground truth density map ($\mathbf{GT}$) and blur each head point $\mathbf{h}_p = (x,y)$ in an image with a Gaussian kernel ($\sigma$). 

\subsection{Scale-aware soft attention masks} \label{sec:sel}

Our approach enriches the baseline with $C$ density estimates $\mathbf{D} = [\mathbf{D}_1, \ldots, \mathbf{D}_C]$, with the idea that each map will be specialized to perform well on a specific range of head sizes. Our network estimates all density maps in a single forward pass by branching the features from intermediate layers of our backbone and sending each into its own regression head.
Then, to aggregate these density estimates and produce a single density estimate $\mathbf{D^F}$, we use a soft attention mechanism that learns a set of $C$ gating masks $\mathbf{M} = [\mathbf{M}_1, \ldots, \mathbf{M}_C]$ corresponding to each branch. Each mask is used to re-weight the pixels of its corresponding density estimate to produce the final density estimate $\mathbf{D^F}$ as follows:
\vspace{-3mm}
\begin{equation}
\begin{aligned}
  & \mathbf{D^F} = \sum_c^C{\mathbf{M}_c \odot \mathbf{D}_c} \\
 \end{aligned}
 \label{eq:scale_aggregation}
\end{equation}
where $\odot$ refers to the element-wise product. The $C$ attention masks are generated by the attention block, which takes as input the last feature map from our backbone network, passes it through an attention head and produces $C$-channel logit maps $\mathbf{Z} = [\mathbf{Z}_1, \ldots, \mathbf{Z}_C]$.  These are then fed to a softmax layer to produce the masks:
\begin{equation}
\mathbf{M}_c^{(i,j)} = \frac{exp( \mathbf{Z}_c^{(i,j)})}{\sum_c exp( \mathbf{Z}_c^{(i,j)})}
\label{eq:mask}
\end{equation}
where $\mathbf{M}_c^{(i,j)}$ and $\mathbf{Z}_c^{(i,j)}$ are the values for the corresponding maps at pixel location $(i,j)$. The softmax ensures that the attentions maps act as a weighted average over the density predictions.

We train this network end-to-end with the same loss used in sec.~\ref{sec:baseline} applied only to the final density estimate ($\mathbf{D^F}$). The intuition is that the attention masks ($\mathbf{M}$) will learn to attend to large heads from the density map predicted by the last branch ($\mathbf{D}_C$) as it is derived from a feature map with a large receptive field. Conversely, smaller heads will be attended by density estimates from earlier branches ($[\mathbf{D}_1, \cdots, \mathbf{D}_{C-1}]$) as these branches have smaller receptive fields and higher spatial resolution, thus they capture finer details in the image.

\subsection{Scale-aware loss regularization} \label{sec:scaleawareloss}
In eq.~\ref{eq:scale_aggregation}, the error signal propagated back to the $c$-th branch is modulated by the attention mask $\mm_c$, i.e., $\frac{\partial \mathcal{L}_{\ell_2}(\mathbf{D^F}, \mathbf{GT})}{\partial \mathbf{D^F}} \odot \mm_c$. Instead of propagating the whole error signals back to every branch, these masks force each branch to only focus on improving the crowd counting accuracy on some selected areas. While fig.~\ref{fig:visualization_attention_map} shows that the attention masks mostly attend to heads of different sizes, as intended, the network has no explicit regularization that enforces this to happen.

Here we present a new scale-aware loss function to further regularize each branch estimate and guide them to specialize on a particular head size. 
To achieve this, we add a scale-aware $\mathcal{L}^{sc}$ loss to each branch, which measures the distance between the branch's predicted density map and our ground truth density map only in areas of the image with heads in a target size range for that density map. In this way, each branch only needs to perform well on its scale.

For each ground truth head point $\mathbf{h}_p$ we estimate the head size $\eta(\mathbf{h}_p)$ and assign it to one of $C$ head size bins. 
The method for predicting $\eta(\mathbf{h}_p)$ is described in sec.~\ref{sec:scale_data_collection}.
To generate each scale supervision mask $\mathbf{S}_c \in \{\mathbf{S}_1, \ldots, \mathbf{S}_C\}$, we set to 1 the regions $\sigma \times \sigma $ around each training head $\mathbf{h}_p$ assigned to bin $c$. 

This supervision guides each map to correctly predict the heads at its scale range, but it does not give any penalty to heads outside of it.
We compute the new scale-aware loss as follows:
\begin{equation}
\mathcal{L}^{sc} =  \frac{1}{WH} \sum_{i}^{W} \sum_{j}^{H} \mathbf{S}_c^{(i,j)}\norm{\mathbf{D}_{c}^{(i,j)} - \mathbf{GT}^{(i,j)}}_{2}^{2}
\label{eq:sal}
\end{equation}

Finally, our final loss is the combination of the ${\ell_2}$ loss on the final density (eq.~\ref{eq:l2loss}) and our scale-aware losses on the intermediate layers of the network.

\begin{equation}
\begin{aligned}
   \mathcal{L} = \mathcal{L}_{\ell_2}(\mathbf{D^F},\mathbf{GT}) + \lambda \sum_{c}^{C} \mathcal{L}^{sc}(\mathbf{D}_c,\mathbf{GT}, \mathbf{S}_c)
 \end{aligned}
 \label{eq:final_loss}
\end{equation}
where $\lambda$ refers to the regularization weight.

\subsection{Estimating the size of each head: $\eta(\mathbf{h}_p)$}
\label{sec:scale_data_collection}

The scale-aware loss regularization presented in the previous section requires an estimate of the diameter of the head $\eta(\mathbf{h}_p)$, however, head size is not available in any crowd counting dataset. In this section we present a new method to estimate it.
We combine the popular geometry-adaptive technique $\eta_{GA}$~\cite{zhang16cvpr} with a new bounding-box-adaptive technique $\eta_{BB}$ that estimates head sizes based on the output of a head detector. More specifically, given head $\mathbf{h}_p$, we estimate its size as follows:
\begin{equation}
\eta(\mathbf{h}_p) = \min(\eta_{GA}(\mathbf{h}_p), \eta_{BB}(\mathbf{h}_p))
\end{equation}

We compute $\eta_{BB}$ by first running a person head detector. Then, for each ground truth head point, we estimate its scale as the median size prediction from the $k$ nearest head detections:
\vspace{-3mm}
\begin{equation}
\eta_{BB}(\mathbf{h}_p) = \text{median}_{z\in kNN_{bb}(\mathbf{h}_p)}\max(w_{z},h_{z})
\label{eq:sigma_bb}
\end{equation}
where $kNN_{bb}(\mathbf{h}_p)$ are the $k$ detected bounding boxes with the closest center to $\mathbf{h}_p$ and $w_z$ and $h_z$ are the width and height of bounding box $z$ respectively. This estimate is only as good as the detector. We found that our detector works well most of the time, but it fails when people are too small and too close together.
Thus, we augment this prediction with the geometry-adaptive approach ($\eta_{GA}(\mathbf{h}_p)$) from Zhang et al. \cite{zhang16cvpr}. For each head, this measure is computed as half the mean distance to the $k$ nearest heads or:
\begin{equation}
\eta_{GA}(\mathbf{h}_p) = \frac{1}{k}\sum_{j\in kNN_{\ell_2}(\mathbf{h}_p)} \sqrt{(\mathbf{h}_p-\mathbf{h}_j)^2}
\label{eq:sigma_ga}
\end{equation}
where $k$ is the number of neighbors and $\mathbf{h}_p$ is the $(x,y)$ location of the ground truth head annotation. This measure works well for crowded scenes but not when people are further apart, thus complementing our $\eta_{BB}$ measure well.

\section{Experiments} \label{sec:exp}

\subsection{Evaluation metrics}
In crowd counting, the count accuracy is measured by two error metrics: Mean Absolute Error (MAE) and Mean Squared Error (MSE), which are defined as follows:

\begin{equation}
MAE = \frac{1}{N} \sum_{i=1}^{N}|C_{i}^{pred} - C_i^{gt}|
\end{equation}
\begin{equation}
MSE = \sqrt{\frac{1}{N} \sum_{i=1}^{N}|C_{i}^{pred} - C_i^{gt}|^2},
\end{equation}

where $N$ is the number of test images, $C_{i}^{pred}$ the predicted count for image $I_i$ and $C_{i}^{gt}$ the ground-truth. 

\subsection{Datasets} \label{sec:datasets}
We evaluate our approach on 4 publicly available crowd counting datasets:
UCF-QNRF~\cite{idrees18eccv}, ShanghaiTech A \& B~\cite{zhang16cvpr} and UCF\_CC\_50~\cite{idrees13cvpr}.

\vspace{1mm}
\noindent {\bf UCF-QNRF} (2018) is the latest released dataset and it consists of 1535 challenging images from Flickr, Web Search and Hajj footage. The number of people in an image (i.e., the count) varies from 49 to 12,865, making this the dataset with the largest crowd variation. Furthermore, the average image resolution is larger compared to all other datasets, causing the absolute size of a person's head to vary drastically from a few pixels to more than 1500. 

\vspace{1mm}
\noindent {\bf ShanghaiTech} (2016) consists of two parts: A and B. Part A contains 482 images of dense scenes like stadiums and parades; its count varies from 33 to 3139. Part B contains 716 images of street scenes from fixed cameras capturing sparser crowds; its count varies from 12 to 578.

\vspace{1mm}
\noindent {\bf UCF\_CC\_50} (2013) consists of 50 black and white, low resolution images and its count varies from 94 to 4543. We follow the dataset instructions and evaluate our results using 5-fold cross-validation. 

\begin{figure*}
\begin{center}
		\includegraphics[width=\textwidth]{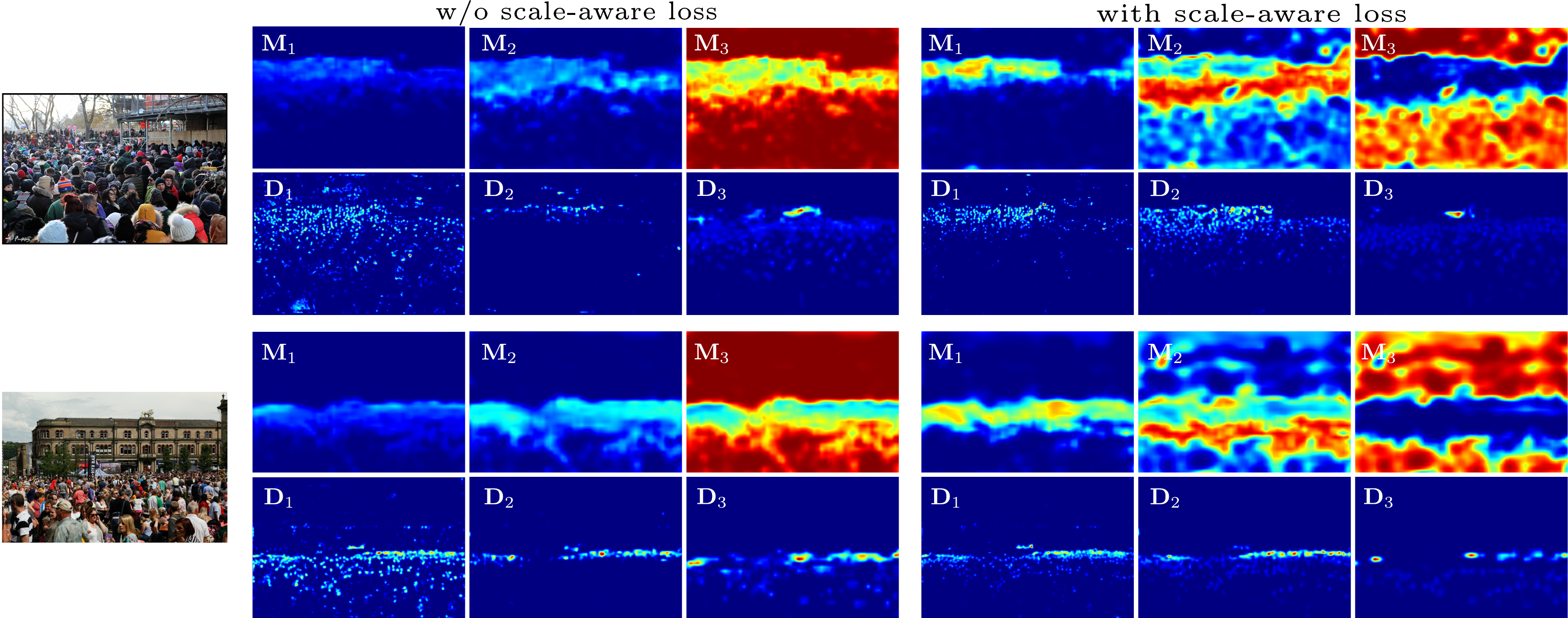}
	\end{center}
	\vspace{-3mm}
\caption{\it We show the attention masks and their corresponding density maps from branches 1,2,3. In general, branch 1 attends to small people (the crowd), while branch 3 attends to larger ones and background regions. After scale-aware loss regularization is introduced, $\mm_1$ and $\mm_2$ get higher weights for small and medium-size people respectively.} \vspace{-3mm}
\label{fig:visualization_attention_map}
\end{figure*}

\subsection{Implementation details} \label{sec:id}

\noindent {\bf Network architecture.}
Similar to other recent crowd counting works~\cite{boominathan16acm,sam17cvpr,sindagi17iccv,li18cvpr}, we use a VGG-16 backbone~\cite{simonyan15iclr}.
We use three branches ($C=3$) from VGG features conv3\_3, conv4\_3 and conv5\_3 from blocks 3, 4 and 5, respectively. 
Our regression head consists of two $3\times3$ convolutions with 128 and 64 channels each, followed by a final $1\times1$ convolutional regression layer.\\

\noindent {\bf Network hyper-parameters.}
We split the training set of the UCF-QRNF dataset into train (80\%) and validation (20\%) and chose the optimal model hyper-parameters on the validation set. We repeated this process on ShanghaiTech and, interestingly, we observed the optimal parameters to be very similar to those for UCF-QRNF. 
We initialized all the new layers in our network with random weights drawn from a Gaussian distributions with zero mean and a standard deviation of 0.0033. We used Adam optimizer \cite{kingma14iclr} and trained our networks for 120 epochs with an initial learning rate of 1e-4, which drops to 1e-5 after 80 epochs. We used a batch size of 64 and input to the network crops of size $384 \times 384$ randomly sampled from different locations in each training image. Following previous works \cite{zhang16cvpr}, we set $\sigma=15$.
To select our head size bins, we split all training heads by size (estimated as described in sec.~\ref{sec:scale_data_collection}) into three buckets with roughly the same number of training examples. This ensures that each intermediate layer has sufficient training data. In the supplementary material (fig. 1), we show visual examples of scale supervision masks generated by binning the head sizes. Finally, we set lambda for eq.~\ref{eq:final_loss} to be 0.1, which provides a good balance between only relying on the loss function $\mathcal{L}_{\ell_2}$ ($\lambda=0$) and only relying on the intermediate losses $\mathcal{L}^{sc}$ ($\lambda>>1$). \\

\noindent {\bf Head estimation.} For both the geometry-adaptive and bounding-box-adaptive estimations, we set the number of neighbors to $3$.
For the bounding-box adaptive estimation, we trained a Faster-RCNN~\cite{ren15nips} head detector with a ResNet-50 backbone~\cite{he16cvpr}. We used the same hyper-parameters as \cite{girshick18detectron}, but we reduced the smallest anchor box size from 32 pixels to 8, in order to be able to localize extremely small heads. We trained our detector on the combination of two public datasets: SCUT-HEAD~\cite{peng18icpr} and Pascal-Parts~\cite{chen14cvpr}. SCUT-HEAD contains annotations for around 111k heads, which are visually similar to those in crowd counting images. Pascal-Parts, on the other hand, contains annotations for only 7.5k heads, but it offers a large selection of extremely useful and difficult background. We found the combination of these two complementary datasets to lead to great detection performance (fig.~\ref{fig:scale_estimation}).

\subsection{Analysis of our method} \label{sec:study}
In this section we explore some of our model components and analyze their outputs. We conduct all experiments on the UCF-QNRF dataset, as it is the largest both in number of images and diversity of crowd count s(sec.~\ref{sec:datasets}).

\vspace{1mm}
\noindent {\bf Multi-scale density ($\mathbf{D}$).}
Li et. al. ~\cite{li18cvpr} showed that the three columns of MCNN~\cite{zhang16cvpr} learned  similar information instead of being specialized to a specific scale.
Here we investigate the predictions of our branches and to what extent our branches are learning different scale information (fig.~\ref{fig:visualization_attention_map}). 
Interestingly, even without our scale-aware loss, our approach is capable of learning different scale information: branch 1 (output $\mathbf{D_1}$) has stronger activations on smaller people, as it relates small-size people with low-level texture patterns, while branch 2 and 3 make far fewer errors on medium and large people, as they operate on a larger receptive field. This complementarity in scale information is essential towards better scale-aware representations. 

\vspace{1mm}
\noindent {\bf Attention mask ($\mathbf{M}$).} In fig.~\ref{fig:visualization_attention_map} we also show the attention masks generated by our approach. Our network learns distinctive attention masks for each branch.  In general, $\mm_3$  has higher weights for large-size people and background regions, while $\mm_1$ gives higher weights to small-scale people.
Interestingly, without our scale aware-loss regularization, the masks learn to attend mostly to the density map predicted by branch 3 (i.e., red region in $\mathbf{M_3}$). On the other hand, when scale-aware loss is used,  $\mm_1$ and $\mm_2$ get higher weights for small and medium-size people, respectively. This demonstrates the importance of using our regularization to help the branches better specialize to their respective scales.

\vspace{1mm}
\noindent {\bf Aggregating multi-scale maps.} We compare our soft attention mechanism (sec.~\ref{sec:sel}) used to aggregate our multi-scale density predictions against other popular aggregation methods: `average', which is popular in semantic segmantation~\cite{long15cvpr}, `max', which is popular in human pose estimation~\cite{cao17cvpr} and `concatenation+conv', which has been used in several multi-column works for crowd counting~\cite{zhang16cvpr,sindagi17iccv,zhang18wacv}. Results are presented in table~\ref{table:pooling}. `Max' produces the largest error, as it tends to over-estimate the count; `concatenation+conv' and `average' work better, but the best performing approach is our attention mechanism. This result proves the effectiveness of our scale-aware aggregation mechanism to fuse multi-scale density maps.

\begin{table}
	\centering
	\small
	\begin{tabular}{l | c c }
		\hline
		Aggregation method & MAE & MSE   \\ \hline
		Average & 123.2 & 206.9 \\
		Max & 144.6 & 227.9 \\
		Concatenation + Conv & 128.3 & 210.1 \\
		Attention (Ours) &  \textbf{116.7} & \textbf{184.5} \\\hline
	\end{tabular}
	\vspace{1mm}
	\caption{\it Evaluation of different pooling techniques to combine the different branches of the network.\vspace{-3mm}}
	\label{table:pooling}
\end{table}

\begin{figure*}
	\begin{center}
		\includegraphics[width=0.9\textwidth]{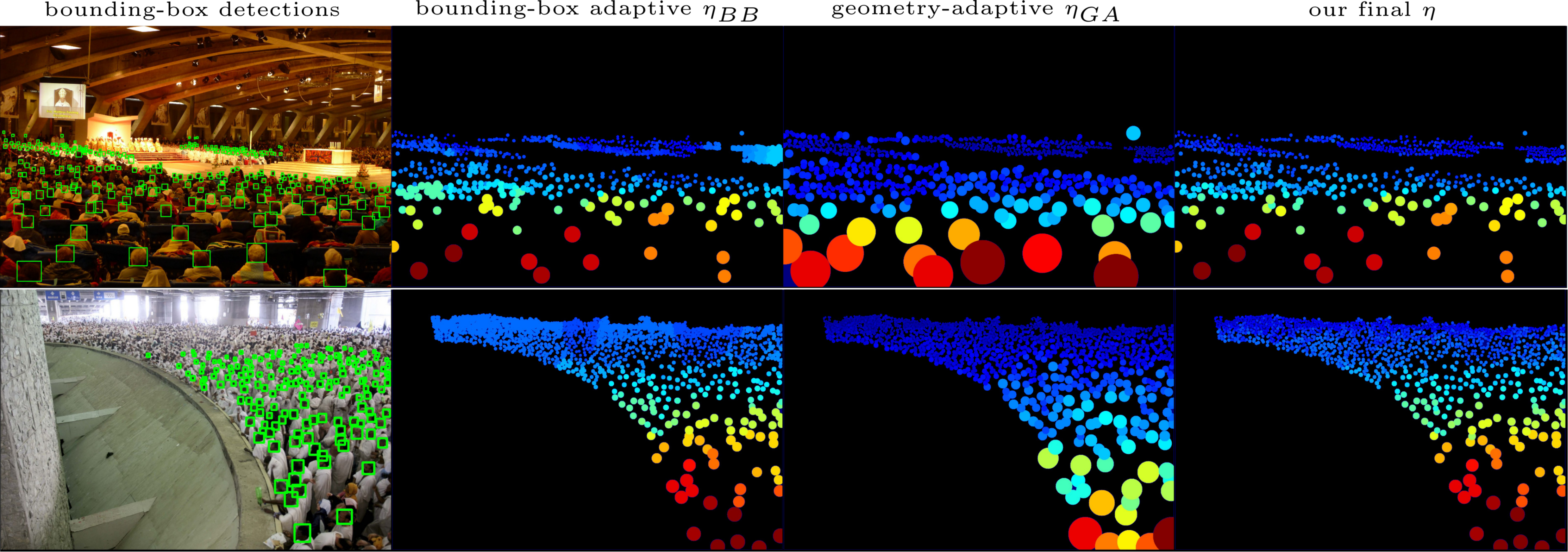}
	\end{center}
	\vspace{-3mm}
	\caption{\small \it Head size estimation. Each map is normalized independently based on its largest (dark red) and smallest head (dark blue). $\eta_{GA}$ tends to perform poorly in sparse regions where faces are far from each other, while $\eta_{BB}$ tends to predict slightly larger sizes than in reality on undetected heads. By combining these two signals, we are able to produce very accurate density maps ($\eta$).}
	\label{fig:scale_estimation}
\end{figure*}

\vspace{1mm}
\noindent {\bf Our head size estimation approach.} 
We present some visual results (fig.~\ref{fig:scale_estimation}) for our head size estimation approach (sec.~\ref{sec:scale_data_collection}). The figure shows two images with the bounding boxes detected by our head detector and the corresponding head sizes estimated by the popular $\eta_{GA}$, our $\eta_{BB}$ and our final $\eta$. For visualization purposes, we color each head based on its size, where dark red is used for the largest head in each map and dark blue for the smallest. $\eta_{GA}$ performs relatively well on very crowded scenes (fig.~\ref{fig:scale_estimation}, bottom row), but it performs rather poorly on sparse regions with small heads far away from each other (fig.~\ref{fig:scale_estimation}, top row). This is probably the reason why CSRNet~\cite{li18cvpr}, among other works, uses the geometry-adaptive Gaussian for the extremely dense ShanghaiTech Part A dataset, but a fixed Gaussian for all other crowd counting datasets.
On the other hand, $\eta_{BB}$ performs very well on both sparse and dense scenes, but it tends to predict slightly larger sizes than in reality on undetected heads (fig.~\ref{fig:scale_estimation}, bottom image, end of the tail).
By combining these two techniques in the novel $\eta$, we are able to overcome most of their limitations and to produce highly accurate maps (fig.~\ref{fig:scale_estimation}, last column).

We now quantitatively compare the performance of our scale-aware approach trained with head binnings defined by (i) our estimation $\eta$ and (ii) the classic geometry-adaptive technique $\eta_{GA}$. While both models achieve comparable MAE performance (around 97.5), the results show a difference in MSE (175.3 vs 167.8, in favor of using our estimate). This is caused by the geometry-adaptive estimate making larger mistakes on the images that contain sparse crowds with tiny heads. As these heads get binned with larger ones, they get assigned to the wrong branch, which now needs to predict heads at a scale outside of its domain (fig.~\ref{fig:network_architecture}), effectively making the training data noisy. 

Finally, we would like to point out that the utility of our head estimation technique goes beyond what we propose in this paper and it can open new research directions for future crowd counting ideas. 

\subsection{Ablation study}
In this section we present incremental results of our model and its components (sec.~\ref{sec:method}). Again, we use the UCF-QNRF dataset for these experiments.\\

\noindent {\bf Baseline results.} As a baseline, we train a VGG-16 architecture backbone with the $\mathcal{L}_{\ell_2}$ loss of eq.~\ref{eq:l2loss} and the settings of sec.~\ref{sec:baseline}. This generates a single feature map from the last convolutional layer of the network and it achieves an MAE of 128.5 (table \ref{table:comp}, row 1) which is the highest error across all entries in the table. Still, this simple baseline achieves competitive results, which is comparable to the state-of-the-art (table \ref{table:results}).
\vspace{1mm}
 
\noindent {\bf Adding $\mathbf{M}$.} By enriching the baseline with three branches that predict multi-scale density maps and our novel scale-aware attention mechanism (sec.~\ref{sec:sel}), the error decreases to 116.7 (table \ref{table:comp} row 3), which is a significant improvement. This indicates that (i), using multi-scale feature maps is beneficial for crowd counting and (ii), the inferred attention masks are performing well on aggregating multi-scale predictions from our multi-branch network.

\vspace{1mm}
\noindent {\bf Adding $\mathcal{L}^{sc}$.} Adding our scale-aware loss regularization (sec.~\ref{sec:scaleawareloss}) also brings an improvement and the error further decreases to 113.3 (table \ref{table:comp}, row 3). This indicates that our regularization helps each branch output accurate density maps for the people within its assigned scale range, which collectively  contributes to the the accuracy improvement of crowd estimation. 

\vspace{1mm}
\noindent {\bf Adding image resolution regularization.} While scale variations can be learned during training, sometimes these exceed the capability of the network. We observed this to be the case for the UCF-QNRF dataset: some of its images are 6k$\times$9k and they contain heads of 1.5k$\times$1.5k, which, clearly, are outside the range of our network's receptive field. To overcome this resolution issue, at inference we down-sample these large images to a maximum size of 1080p. This simple normalization improves performance considerably for all components, lowering our full approach MAE to 97.5 (table \ref{table:comp}, last row). It is also worth noticing that while this improvement is generic, it still preserves the relative importance of each model component: adding $\mathbf{M}$ still improves substantially over the baseline and adding $\mathcal{L}^{sc}$ still remains the best performing model. Interestingly, adding $\mathcal{L}^{sc}$ improves MSE considerably over just $\mathbf{M}$ (175.6 to 167.8).

\begin{table}
\centering
\resizebox{0.85\columnwidth}{!}{
	\begin{tabular}{c : c : c : c | c c }
		\hline
     VGG16 &  +$\mathbf{M}$ &  +$\mathcal{L}^{sc}$ & +ImgRes & MAE & MSE\\\hline
	 \cmark & & & & 128.5 & 205.6 \\
 	 \cmark & \cmark & & & 116.7 & 184.5 \\
  	 \cmark & \cmark & \cmark & & 113.3 & 183.2 \\\hline
  	 \cmark &  &  & \cmark & 109.7 & 186.5 \\
  	 \cmark & \cmark &  & \cmark & 99.6 & 175.6 \\
	 \cmark & \cmark & \cmark & \cmark & {\bf 97.5} & {\bf 167.8} \\\hline
	\end{tabular}}
	\vspace{0.5mm}
	\caption{\it Results for the different components of our architecture.}
	\label{table:comp}
\end{table}

\begin{table*}
\centering
\resizebox{1.8\columnwidth}{!}{
  \begin{tabular}{l | r l | c c | c c | c c | c c }
    \hline
    \multicolumn{3}{c|}{} & \multicolumn{2}{c|}{UCF-QNRF} & \multicolumn{2}{c|}{ShanghaiTechA} & \multicolumn{2}{c|}{ShanghaiTechB} & \multicolumn{2}{c}{UCF\_CC\_50}\\
     \textbf{Method} & \multicolumn{2}{c|}{\textbf{Venue \& Year}} & MAE   & MSE   & MAE   & MSE   & MAE   & MSE   & MAE & MSE \\ \hline
     MCNN~\cite{zhang16cvpr}    & CVPR & 2016	& 277 	& 426 	& 110.2 & 173.2 & 26.4 	& 41.3  & 377.6 & 509.1 \\
    C-MTL~\cite{sindagi17avss}  & AVSS & 2017 	& 252 	& 514 	& 101.3 & 152.4 & 20.0 	& 31.1   & 322.8 & 341.4 \\
    SwitchCNN~\cite{sam17cvpr}  & CVPR & 2017	& 228 	& 445 	& 90.4 	& 135.0 & 21.6 	& 33.4   & 318.1 & 439.2 \\
    CP-CNN~\cite{sindagi17iccv} & ICCV & 2017  & - 	& - 	& 73.6 	& 106.4 & 20.1 	& 30.1  & 295.8 & \bf 320.9 \\
    SaCNN~\cite{zhang18wacv}    & WACV & 2018  & -     & -     & 86.8  & 139.2 & 16.2  & 25.8  & 314.9 & 424.8 \\
	ACSCP~\cite{shen2018cvpr}   & CVPR & 2018  & -     & -     & 75.7  & 102.7 & 17.2  & 27.4  & 291.0 & 404.6 \\
	IG-CNN~\cite{sam18cvpr}     & CVPR & 2018  & -     & -     & 72.5  & 118.2 & 13.6  & 21.1 & 291.4 & 349.4 \\
	Deep-NCL~\cite{shi18cvpr}	& CVPR & 2018  & -     & -     & 73.5  & 112.3 & 18.7  & 26.0 & 288.4 & 404.7 \\
    CSRNet~\cite{li18cvpr} 	    & CVPR & 2018	& - 	& - 	& 68.2 	& 115.0 & 10.6 	& 16.0  & 266.1 & 397.5 \\ 
    CL-CNN~\cite{idrees18eccv}  & ECCV & 2018  & \bf 132 	& \bf 191 	& - 	& - 	& - 	&  -  & - & - \\
    SANet~\cite{cao18eccv}      & ECCV & 2018  & -  	& -	    & \bf 67.0 	& \bf 104.5 & \bf 8.4 	&  \bf 13.6 & \bf 258.4 & 334.9 \\\hline
    Our Baseline (from scratch) & \multicolumn{2}{c|}{-}     & 109.7    & 186.5 &  71.9 & 107.1 &  8.9 	& 14.4  & 267.1 &  351.7\\
    Our Approach (from scratch)     & \multicolumn{2}{c|}{-}     & \cellcolor[gray]{0.9} \textbf{97.5}    \cellcolor[gray]{0.9} & \cellcolor[gray]{0.9} \textbf{167.8} 	& \cellcolor[gray]{0.9} \textbf{65.6} 	& \cellcolor[gray]{0.9} \textbf{102.1} 	&  \cellcolor[gray]{0.9} \textbf{8.3} & \cellcolor[gray]{0.9} \textbf{13.3}  & \cellcolor[gray]{0.9} \textbf{248.7} &  \cellcolor[gray]{0.9} \textbf{327.2} \\
    Our Approach (w/ pre-training)   & \multicolumn{2}{c|}{-}     &  -  & -	& \cellcolor[gray]{0.8} \textbf{62.1} 	& \cellcolor[gray]{0.8}\textbf{98.5} 	&  \cellcolor[gray]{0.8}\textbf{7.6} & \cellcolor[gray]{0.8}\textbf{12.4}  &  \cellcolor[gray]{0.8}\textbf{238.2} &  \cellcolor[gray]{0.8}\textbf{310.8} \\\hline
  \end{tabular}}
  \vspace{1mm}
  \caption{\it Quantitative results of our approach on four public datasets, against several approaches in the literature.\vspace{-1mm}}
\label{table:results}
\end{table*}

\begin{figure*}
	\begin{center}
		\includegraphics[width=0.85\textwidth]{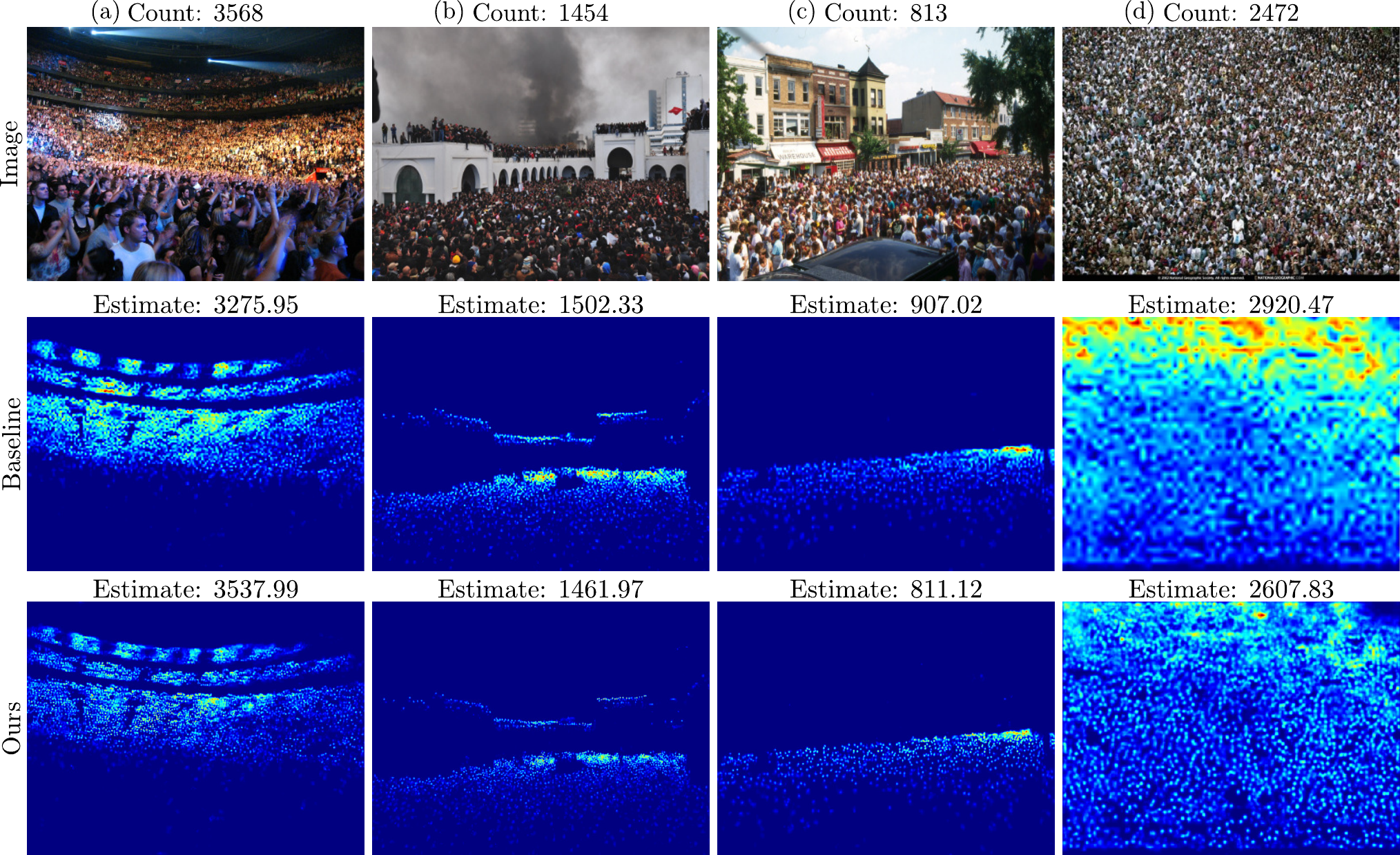}
	\end{center}
	\vspace{-3mm}
	\caption{\small \it Qualitative results. Not only does our approach achieves considerably better count estimates than our baseline, but it also produces sharper and better localized predictions.}
	\label{fig:qualitative_results}
\end{figure*}

\subsection{Comparison to other crowd counting methods}
We now compare our model against several approaches in the literature, on the datasets introduced in sec.~\ref{sec:datasets}.
Results are presented in table~\ref{table:results} and fig.~\ref{fig:qualitative_results}.
As most of the other works in the literature, we train our models from scratch and only use the image resolution regularization on the UCF-QNRF dataset (the only one with extremely large images). 
Overall, our approach always performs better than the baseline, showing the importance of learning multi-scale features. %
Furthermore, our approach also outperforms all previous methods in the literature, on all datasets and all metrics.
Interestingly, we observe the biggest improvement on the UCF-QNRF dataset, which is the largest dataset and the one with the most diverse variation of head sizes. This further shows that our model is capable of handling such large scale variations to produce positive results (fig.~\ref{fig:qualitative_results}).

Additionally, we also present results for our models pre-trained on the large UCF-QNRF dataset. As this dataset was only recently released, no previous work have addressed the impact of pre-training for crowd counting. Our results show that pre-training is very important and it further improves our results by 5-10\% on all datasets.

Finally, fig.~\ref{fig:qualitative_results} presents some visual results of our baseline and our approach. In addition to performing better on counting the number of people in an image, our approach also shows better localized predictions. Its density maps are much sharper than those outputted by the baseline, which tends to oversmooth regions with large crowds. This is especially evident in fig.~\ref{fig:qualitative_results}{\color{red}d}.  
It also validates our hypothesis that directly using low-level layers to output intermediate density maps is beneficial for localizing small-scale people, as these low-level feature maps have detailed spatial layout.

\section{Conclusions} \label{sec:concl}

In this work, we proposed a novel multi-branch architecture that generates multi-scale density maps from its intermediate layers.
To aggregate these density maps into our final prediction, we developed a new soft attention mechanism that learns a set of gating masks, one for each map. We further introduced a scale-aware loss to guide each branch to specialize on different scale ranges . Finally, we proposed a simple, yet effective technique to estimate the size of each head in an image. Our approach achieved state-of-the-art results on four challenging crowd counting datasets, on all evaluation metrics. 

{
\newpage
\bibliographystyle{unsrt}
\bibliography{egbib}
}

\end{document}